\documentclass[9pt]{styles/IEEEtran}

\newcommand{\asplossubmissionnumber}{323}
\usepackage[normalem]{ulem}

\usepackage{caption}
\usepackage{subcaption}
\usepackage{multirow}
\usepackage[table]{xcolor}
\usepackage{cite}
\usepackage{graphicx}
\usepackage{algorithm}
\usepackage[noend]{algpseudocode}
\usepackage{array}
\usepackage{fixltx2e}
\usepackage[table]{xcolor}
\usepackage{stfloats}
\usepackage{url}

\usepackage{times}
\usepackage{color}
\usepackage{verbatim}
\usepackage{subfiles}
\usepackage{lipsum}
\usepackage{xr}
\usepackage{hyperref}
\usepackage{zref-xr}
\usepackage{blindtext}
\usepackage{tabularx}
\usepackage{etoolbox}
\usepackage[font=small,skip=0pt]{caption}
\usepackage{setspace}
\usepackage{export}
\usepackage{nameref}
\usepackage{inputenc}
\usepackage{mathtools}
\usepackage{esvect}
\usepackage{xspace}	
\usepackage{soul}
\usepackage[export]{adjustbox}
\usepackage{amsmath, amsthm, amssymb}
\graphicspath{{../pdf/}{../jpeg/}}
\DeclareGraphicsExtensions{.pdf,.jpeg,.png}
\makeatletter
\patchcmd{\@classz}
  {\CT@row@color}
  {\oldCT@column@color}
  {}
  {}
\patchcmd{\@classz}
  {\CT@column@color}
  {\CT@row@color}
  {}
  {}
\patchcmd{\@classz}
  {\oldCT@column@color}
  {\CT@column@color}
  {}
  {}
\makeatother

\renewcommand{\baselinestretch}{.98}
\definecolor{fxnote}{rgb}{0.8000,0.0000,0.0000}
\hypersetup{
    colorlinks=true,
    linkcolor=red,
    filecolor=magenta,      
    urlcolor=cyan,
}
\renewcommand{\paragraph}[1]{\textbf{Brief: #1}}

\newcommand{\red}[1]{\textcolor{red}{#1}}

\usepackage{hyperref}
\usepackage{authblk}
\begin{document}
\makeatletter

\def\ps@IEEEtitlepagestyle{
  \def\@oddfoot{\mycopyrightnotice}
  \def\@evenfoot{}
}
\def\mycopyrightnotice{
  {\footnotesize xxx-x-xxxx-xxxx-x/xx/\$31.00~\copyright~2018 IEEE\hfill} 
  \gdef\mycopyrightnotice{}
}

\renewcommand\AB@affilsepx{, \protect\Affilfont}
\makeatother
\title{\fontsize{.7cm}{.9cm}\selectfont
 RoboRun: A Robot Runtime to Exploit Spatial Heterogeneity\vspace{-.5em}}
\author[$\dagger$]{Behzad Boroujerdian}
\author[$\ddagger$]{Radhika Ghosal}
\author[$\ddagger$]{Jonathan Cruz}
\author[$\ddagger$]{Brian Plancher}
\author[$\dagger\ddagger$]{Vijay Janapa Reddi\vspace{-.7em}}
\affil[$\dagger$]{The University of Texas at Austin}
\affil[$\ddagger$]{\vspace{-.5em}Harvard University\vspace{-1.5em}}


\date{}
\IEEEoverridecommandlockouts
\IEEEpubid{\makebox[\columnwidth]{978-1-5386-5541-2/18/\$31.00~\copyright2018 IEEE} \hspace{\columnsep}\makebox[\columnwidth]{ }}
\maketitle
\IEEEpubidadjcol

\begin{abstract}
\label{sec:abstract} 
The limited onboard energy of autonomous mobile robots poses a tremendous challenge for practical deployment. Hence, efficient computing solutions are imperative. A crucial shortcoming of state-of-the-art computing solutions is that they ignore the robot's operating environment heterogeneity and make static, worst-case assumptions. As this heterogeneity impacts the system's computing payload, an optimal system must dynamically capture these changes in the environment and adjust its computational resources accordingly. This paper introduces RoboRun, a mobile-robot runtime that dynamically exploits the compute-environment synergy to improve performance and energy. We implement RoboRun in the Robot Operating System (ROS) and evaluate it on autonomous drones. We compare RoboRun against a state-of-the-art static design and show 4.5X and 4X improvements in mission time and energy, respectively, as well as a 36\% reduction in CPU utilization.

\end{abstract}

\section{Introduction}
\label{sec:introduction} 


Autonomous mobile robots are on the cusp of showing their potential with use cases such as package delivery and surveillance. However, their limited onboard energy is a tremendous challenge. This has been traditionally addressed through the development of faster algorithms to improve robot navigation~\cite{informedRRT}.

In contrast, we show that more efficient onboard computing can also be an effective solution. Prior work established that compute can improve mission time and energy as \textit{the higher the computational power, the faster the robot can process and react to its environment}, enabling more agile/higher velocity navigation. As a robot moves faster, it finishes its mission faster, lowering energy consumption~\cite{boroujerdian2018mavbench}.

However, a crucial shortcoming of state-of-the-art computing solutions is that they make static, worst-case assumptions about the robot's operating environment. Concretely, these solutions ignore the \textit{spatial heterogeneity} of 
the environment (physical world) in which robots navigate. For example, a drone traversing tight aisles of a warehouse requires high-precision navigation to avoid collisions. This is in contrast to when it flies over open skies between warehouses which are almost entirely obstacle-free, and therefore only requiring low precision navigation. By ignoring this heterogeneity, mobile robots are forced to make worst-case decisions to ensure their safety. For example, for the above drone, high-precision is always enforced, regardless of whether it is navigating the tight aisles or open skies. This approach, typical of traditional systems, significantly degrades the overall performance as such permanent high-precision results in permanent low velocity throughout the mission~\cite{boroujerdian2018mavbench, liu2016high}.

To address this issue, we propose \textit{spatial aware computing} to continuously and dynamically adjust computational resources according to the robot's spatial environment. 
To systematically analyze this heterogeneity, we identify four key features: space precision, space volume, space traversal speed, and space visibility. We then both mathematically and experimentally quantify their impact on compute. 

We then present \textit{RoboRun}, a spatial aware middleware that dynamically optimizes the robot navigation pipeline according to the environment's spatial features. RoboRun deploys a governor (Figure~\ref{fig:PPC_pipeline} top) to generate optimal policies for the application layer (Figure~\ref{fig:PPC_pipeline} middle) using our time budgeting algorithm, collected cyber-physical models, and a mathematical solver. 
We introduce precision and volume control operators for each navigation pipeline stage (6 knobs in total) to adjust the computational load according to the policy.

We use Micro Aerial Vehicles (MAVs), drones, as our example robot and compare RoboRun against a state-of-the-art baseline, \textit{spatial oblivious} design, whose parameters are set statically at design time. By dynamically optimizing computational resources, RoboRun improves velocity, mission time, and energy consumption by 5X, 4.5X, and 4X. It also reduces CPU utilization by 36\% freeing up computational resources for higher-level cognitive tasks such as semantic labeling. Our environment sensitivity analysis shows that RoboRun is more sensitive (1.5X vs. 1.1X) to environment heterogeneity, indicating its higher environmental awareness. Finally, RoboRun is better suited for long-distance missions as it is less sensitive (1.3X vs. 2X) to the distance of the goal location.

In summary, we make the following contributions:
\begin{itemize}
\item We quantify the impact of \emph{space heterogeneity} and four of its features on the robot's compute. We then introduce \emph{spatial aware computing}, a framework that exploits space-compute interactions to improve computational latency, mission time, and energy.

\item We implement RoboRun, a \emph{spatial aware middleware}, demonstrating how spatial aware computing can be exploited via the use of operators, profilers, and a governor. 

\item We also show that spatial aware designs can lower the navigation workload's pressure on compute.
Since navigation is a primitive task, this frees up resources for higher-level cognitive tasks.
\end{itemize}

\textbf{Related Work:}

%
%
%
%
\textit{System Design for Robotics:} Prior works have explored individual kernel accelerators with statically set knobs~\cite{ murray2016microarchitecture, suleiman2019navion}, made algorithmic improvements ignoring system implications~\cite{sermanet2009multirange, liu2016high}, developed runtime control targeting algorithmic specific knobs~\cite{zhao2019towards, sudhakarbalancing} or individual kernels~\cite{opp_in, dyn_sw}, and demonstrated the importance of system-environment synergy without exploiting them~\cite{boroujerdian2018mavbench, lin2018architectural}. In this work, we instead take a holistic view of the end-to-end pipeline, study spatial heterogeneity systematically and design a middleware that adapts our compute subsystem online to this heterogeneity by targeting fundamental and inherent physical space characteristics.

\textit{Deadline Assignment:} Prior work has focused on the general deadline assignment problem~\cite{marinca2004analysis}. In contrast, RoboRun specifically formulates and responds to the space-induced deadline.

\section{Spatial Aware Computing}
\label{sec:sptl_computing}

As MAVs operate in a heterogeneous world, they need to leverage a spatial awareness to dynamically optimize for such heterogeneity. 
We identify four different features causing space heterogeneity and quantify their impact on compute. We then detail the spatial aware computing model and its impact on overall system performance.

\subsection{Spatial Heterogeneity in the Real World}

As a MAV flies through the environment, its local physical space varies, changing the inputs sent to and the outputs required from its cyber system. For example, a package delivery drone navigating tight warehouse aisles demands high precision while it experiences no such constraints flying over open skies towards its destination.



A compute subsystem that adjusts to such heterogeneity can modulate its decision rate and improve performance. To understand this, we discuss the interactions between 2 fundamental cyber quantities (latency, deadline) and 4 fundamental spatial heterogeneity features (space volume, space precision, traversal speed, and visibility).

\textit{(Decision) latency/deadline:}
Compute subsystem continuously processes the environment making navigational decisions. Decision latency is the processing time of each sampled input, and decision deadline is the maximum latency that can be tolerated while ensuring a collision-free flight. Thus for safety, decision latency must always be less than the decision deadline.

\begin{figure}[t!]
 \vspace{-13pt}
\centering
     
      \begin{subfigure}{0.24\columnwidth}
    \centering
    \includegraphics[trim=120 0 160 0, clip, width=\columnwidth]{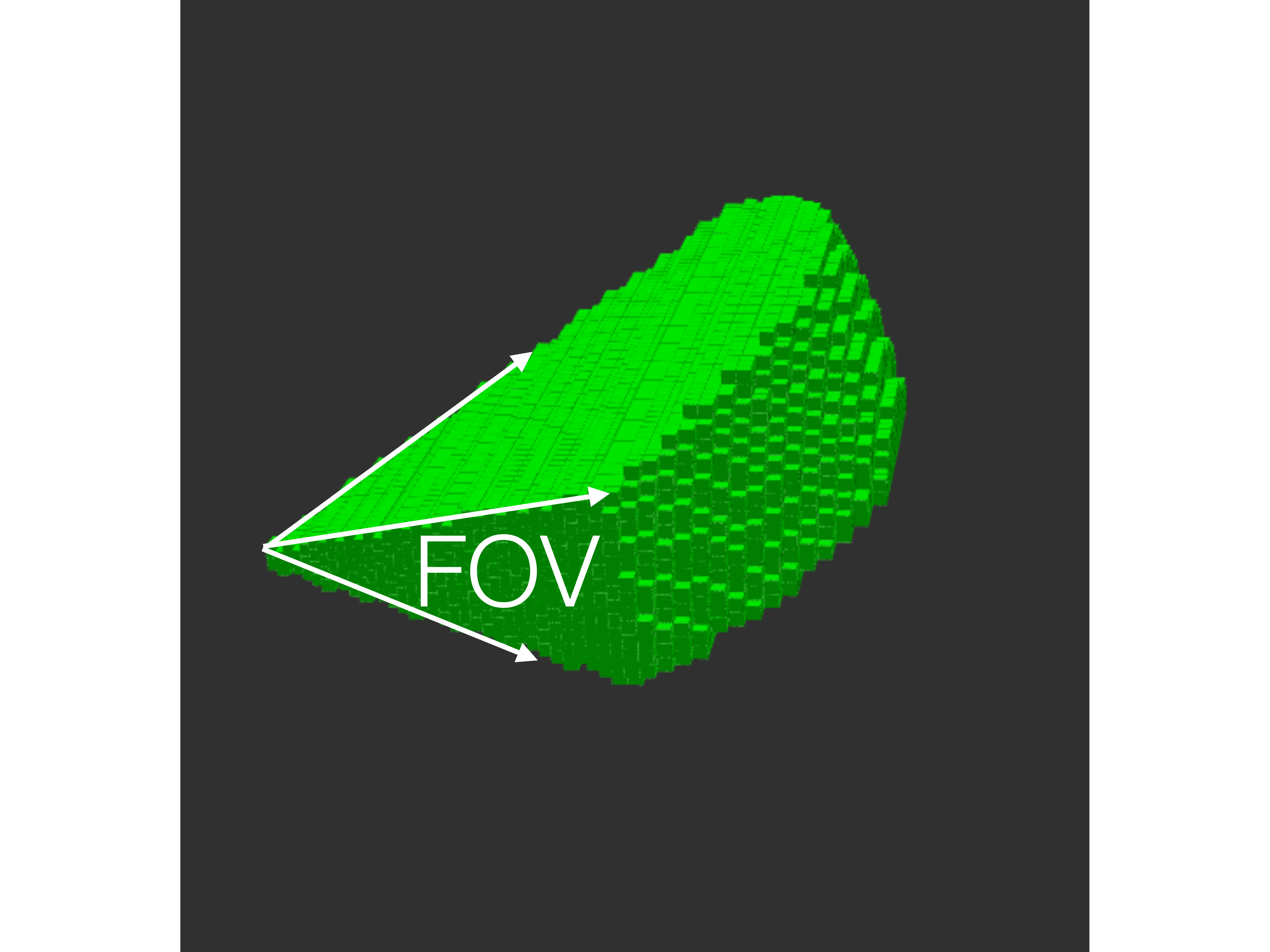}
   \caption{Low volume, fewer voxels.}
    \label{fig:low_volume}
    \end{subfigure}
     \hfill
    \begin{subfigure}{0.24\columnwidth}
    \centering	
    \includegraphics[trim=160 0 120 0, clip, width=\columnwidth]{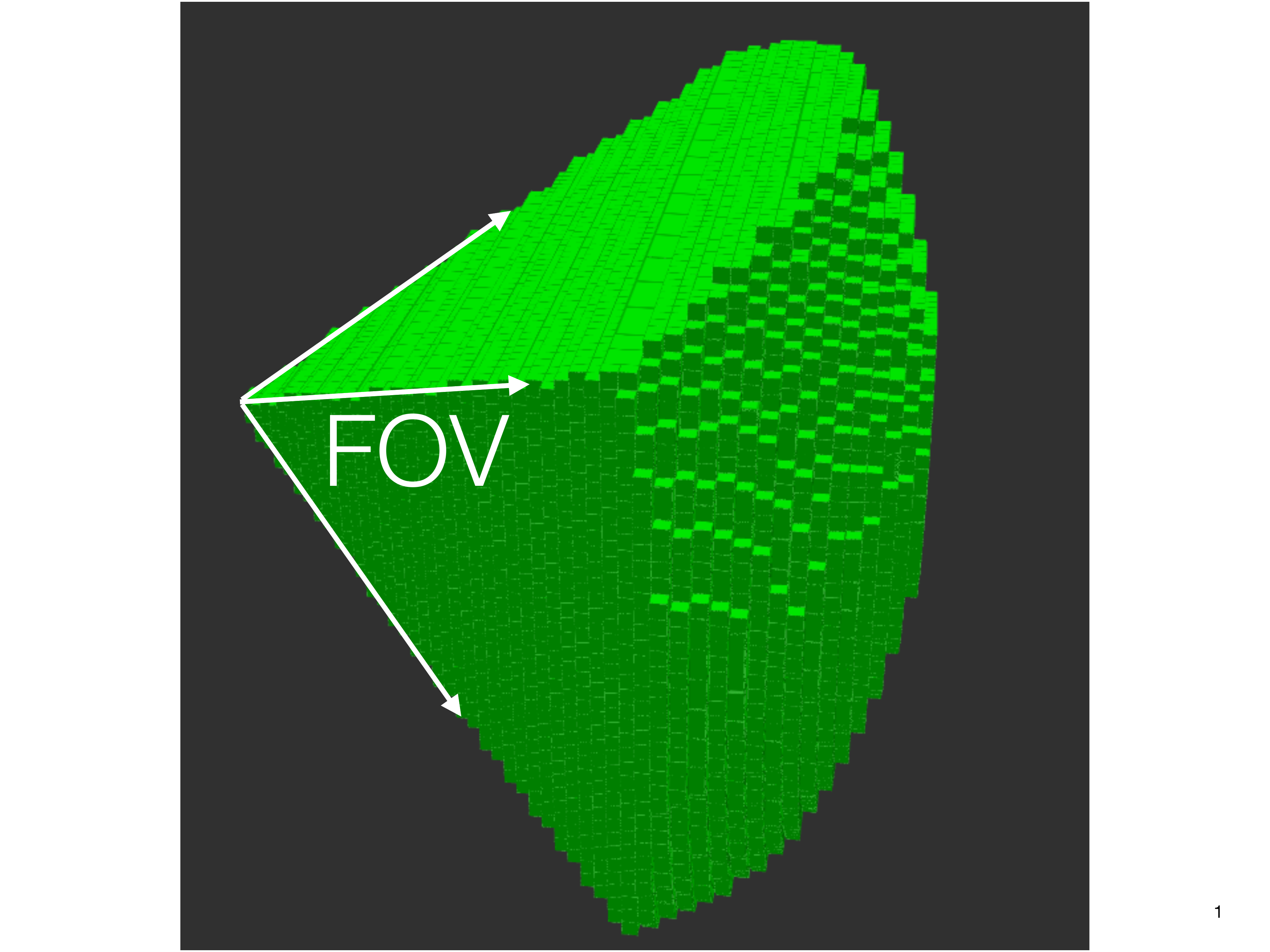}
    \caption{High volume, more voxels.}
    \label{fig:high_volume}
    \end{subfigure}
    \begin{subfigure}{0.24\columnwidth}
    \centering
    \includegraphics[trim=120 0 160 0, clip, width=\columnwidth]{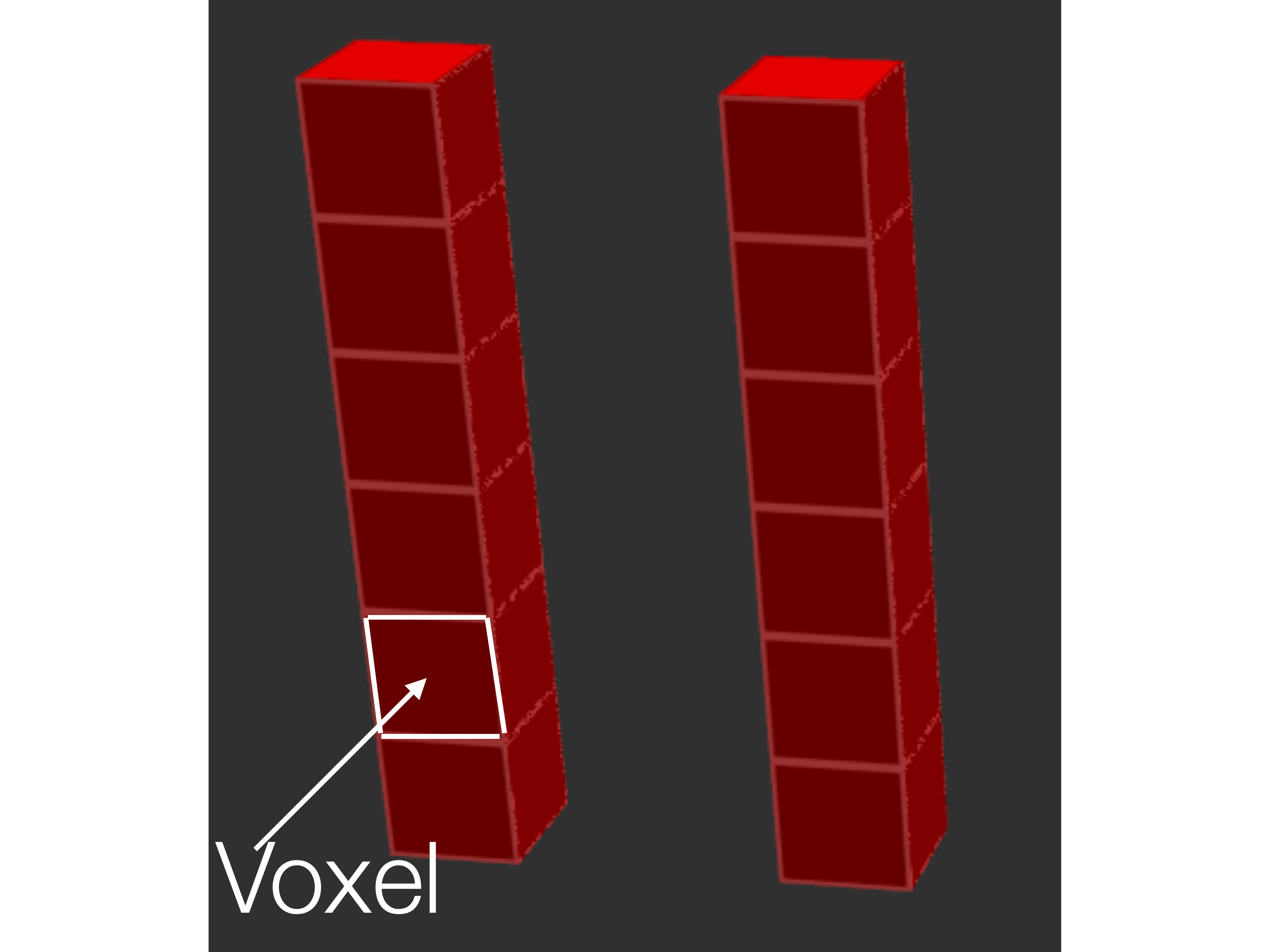}
   \caption{Low precision, fewer voxels.}
    \label{fig:low_precision}
    \end{subfigure}
    \hfill
    \begin{subfigure}{0.24\columnwidth}
    \centering	
    \includegraphics[trim=160 0 120 0, clip, width=\columnwidth]{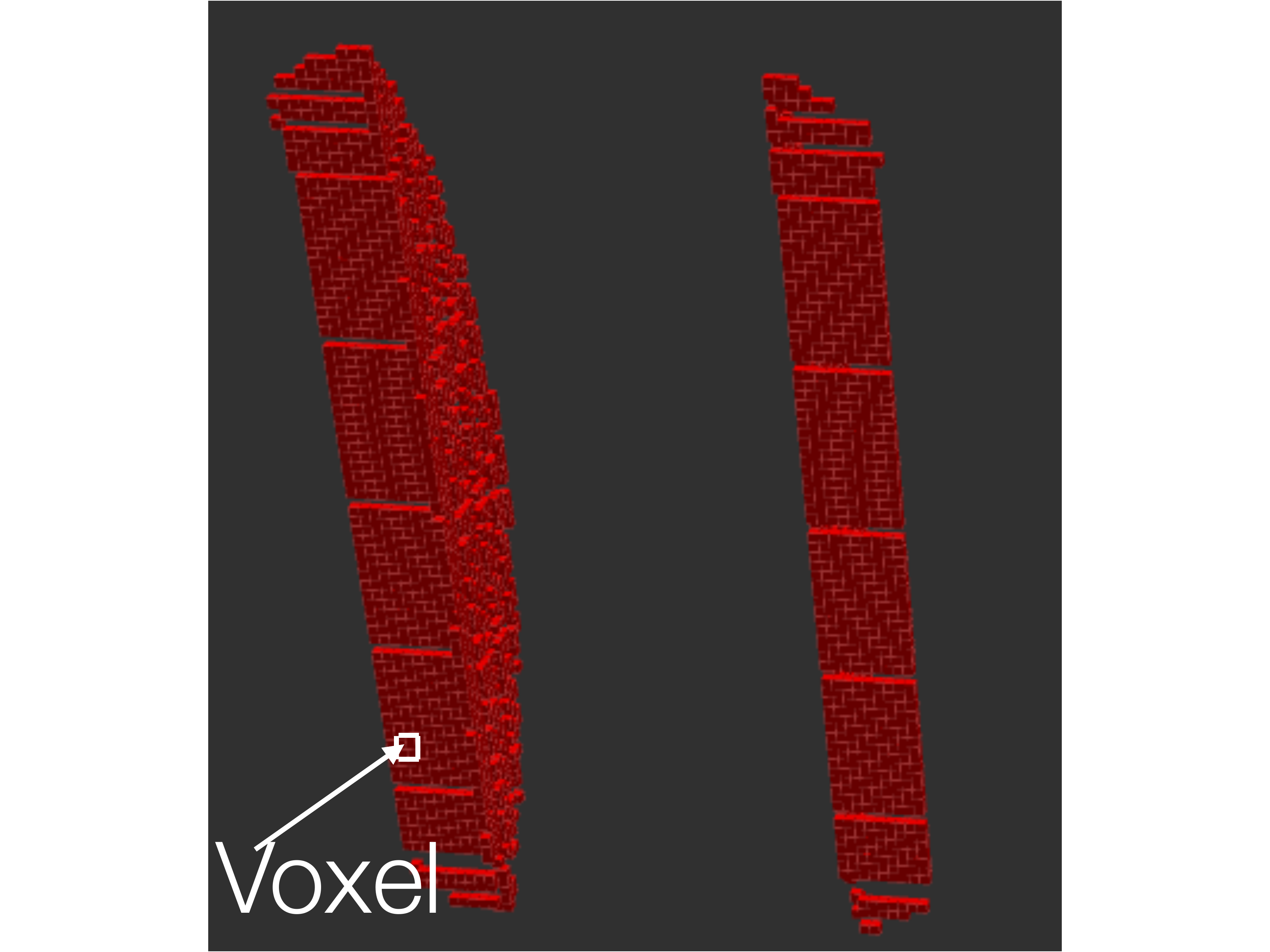}
    \caption{High precision, more voxels.}
    \label{fig:high_precision}
    \end{subfigure}
\caption{Volume/precision impact on number of voxels. FOV stands for field of view.}
\label{fig:dynamic_precision}
\end{figure}

\begin{figure}[t!]
\vspace{-1.95em}
\centering
    \begin{subfigure}{0.49\columnwidth}
    \centering
     \hspace*{-1.5em}
    \includegraphics[trim=5 0 73 0, clip, width=\columnwidth]{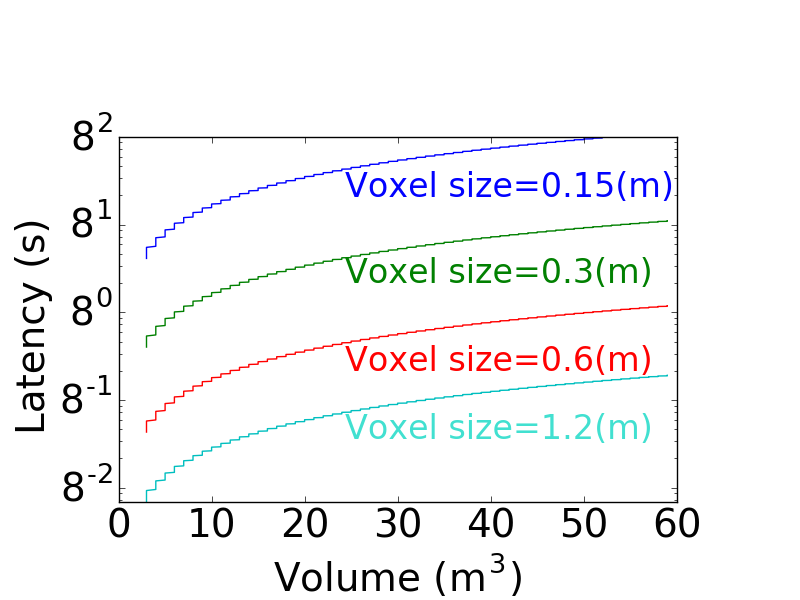}
   \caption{Precision/volume/latency.}
    \label{fig:precision_volume_reponse_time_relationship}
    \end{subfigure}
    \begin{subfigure}{0.49\columnwidth}
    \centering	
    \includegraphics[trim=5 0 75 0, clip, width=\columnwidth]{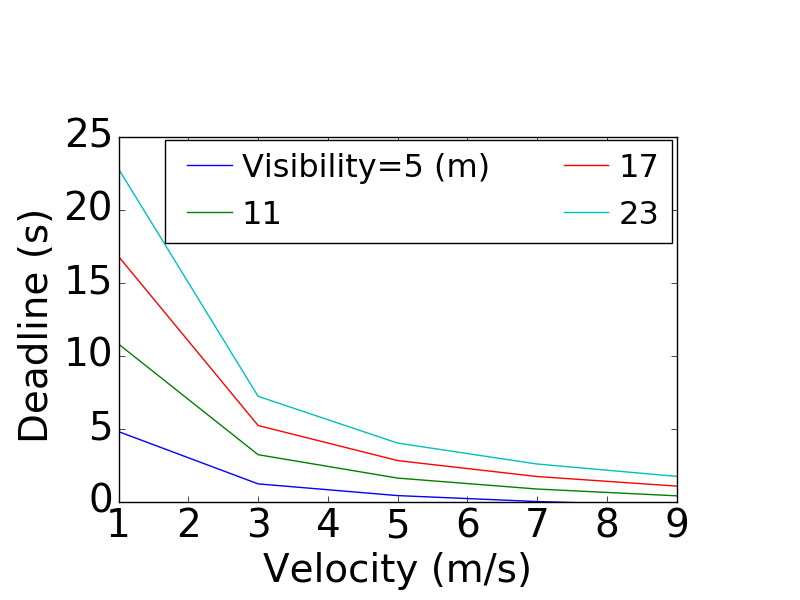}
    \caption{Speed/visibility/deadline.}
    \label{fig:visibility_velocity_time_budget_relationship}
    \end{subfigure}
\caption{Space, processing latency and processing deadline for our MAV.}
\label{fig:dynamic_precision}
\end{figure}

\textit{Space Volume:} The volume of space processed (measured in cubic meters) directly impacts the computational load. Larger volumes require processing more voxels\footnote{Voxel is the smallest unit of 3D space analogous to a pixel in a 2D image.} (Figure~\ref{fig:low_volume},~\ref{fig:high_volume}), increasing the decision latency. A 2X increase in volume requires processing twice as many voxels and hence a 2X increase in latency (Figure~\ref{fig:precision_volume_reponse_time_relationship}).

\textit{Space Precision:} Space precision, the accuracy with which space is processed, directly impacts the computational latency. Increasing precision decreases the size of each voxel (measured in meters). Hence, a more precise map uses smaller yet more voxels to encode the same space (Figure~\ref{fig:low_precision} and~\ref{fig:high_precision}), leading to higher latency. For example, the top curve in Figure~\ref{fig:precision_volume_reponse_time_relationship} with 2X the precision (i.e., half the voxel size) of the second to the top curve has 8X number of voxels, and hence experiences up to an 8X increase in latency. 

\begin{figure*}[!hb]
\centering
    \vspace{-1em}
    \begin{subfigure}{0.15\textwidth}
    \centering
        \includegraphics[valign=t, trim=0 0 0 0, clip, height=2.3pt]{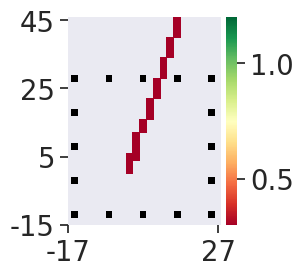}
        \vspace{-0.5em}
        \caption{Precision (m)}
        \label{fig:static_precision}
    \end{subfigure}
    \begin{subfigure}{0.15\textwidth}
    \centering
        \includegraphics[valign=t, trim=0 0 0 0, clip, height=2.3pt]{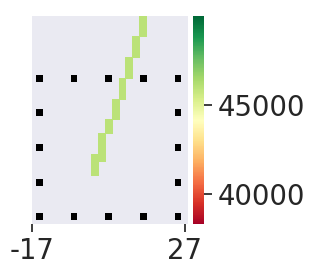}
        \vspace{-0.5em}
        \caption{Volume ($m^3$)}
        \label{fig:static_volume}
    \end{subfigure}
    \begin{subfigure}{0.15\textwidth}
    \centering
        \includegraphics[valign=t, trim=0 0 0 0, clip, height=2.3pt]{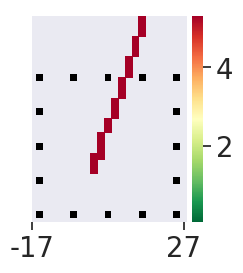}
        \vspace{-0.5em}
        \caption{Latency (s)}
        \label{fig:static_res_vol_resp_time}
    \end{subfigure}
    \begin{subfigure}{0.15\textwidth}
    \centering
        \includegraphics[valign=t, trim=0 0 0 0, clip, height=2.3pt]{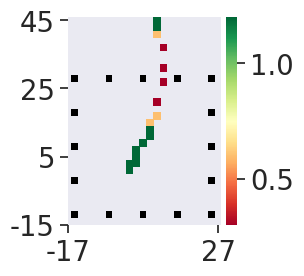}
        \vspace{-0.5em}
        \caption{Precision (m)}
        \label{fig:dynamic_precision}
    \end{subfigure}
    \begin{subfigure}{0.15\textwidth}
    \centering
        \includegraphics[valign=t, trim=0 0 0 0, clip, height=2.3pt]{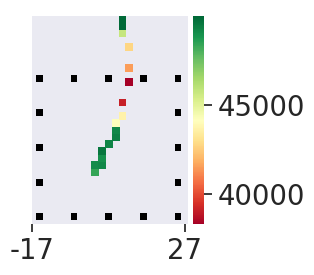}
        \vspace{-0.5em}
        \caption{Volume ($m^3$)}
        \label{fig:dynamic_volume}
    \end{subfigure}
    \begin{subfigure}{0.15\textwidth}
    \centering
        \includegraphics[valign=t, trim=0 0 0 0, clip, height=2.3pt]{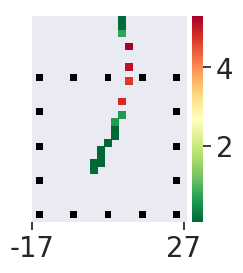}
        \vspace{-0.5em}
        \caption{Latency (s)}
        \label{fig:dynamic_res_vol_resp_time}
    \end{subfigure}
\caption{A high precision mission. Spatial oblivious design (left 3 graphs) uses constant worst-case values comparing to spatial aware design (right 3).}
\label{fig:spatial_aware_design_res_volume}
\end{figure*}

\begin{figure*}[!hb]
    \vspace{-0.8em}
    \centering
        \begin{subfigure}{0.15\textwidth}
        \includegraphics[valign=t, trim=0 0 0 0, clip, height=2.3pt]{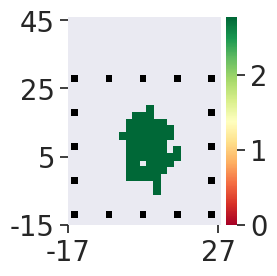}
        \vspace{-0.5em}
        \caption{Velocity (m/s)}
        \label{fig:static_velocity}
    \end{subfigure}
    \begin{subfigure}{0.15\textwidth}
    \centering
        \includegraphics[valign=t, trim=0 0 0 0, clip, height=2.3pt]{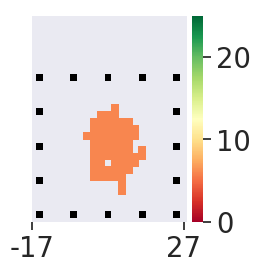}
        \vspace{-0.5em}
        \caption{Visibility (m)}
        \label{fig:static_visibility}
    \end{subfigure}
    \begin{subfigure}{0.15\textwidth}
    \centering
        \includegraphics[valign=t, trim=0 0 0 0, clip, height=2.3pt]{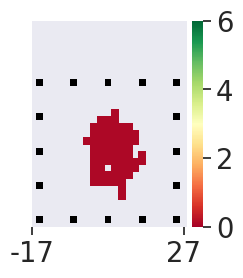}
        \vspace{-0.5em}
        \caption{Deadline (s)}
        \label{fig:static_velocity_budget}
    \end{subfigure}
    \begin{subfigure}{0.15\textwidth}
    \centering
        \includegraphics[valign=t, trim=0 0 0 0, clip, height=2.3pt]{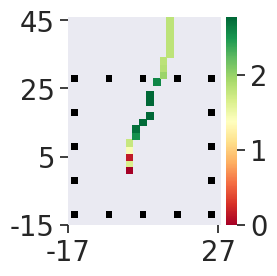}
        \vspace{-0.5em}
        \caption{Velocity (m/s)}
        \label{fig:dynamic_velocity}
    \end{subfigure}
    \begin{subfigure}{0.15\textwidth}
    \centering
        \includegraphics[valign=t, trim=0 0 0 0, clip, height=2.3pt]{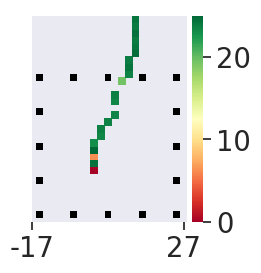}
        \vspace{-0.5em}
        \caption{Visibility (m)}
        \label{fig:dynamic_visibility}
    \end{subfigure}
    \begin{subfigure}{0.15\textwidth}
    \centering
        \includegraphics[valign=t, trim=0 0 0 0, clip, height=2.3pt]{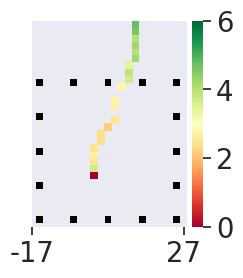}
        \vspace{-0.5em}
        \caption{Deadline (s)}
        \label{fig:dynamic_velocity_budget}
    \end{subfigure}
\caption{A high-velocity mission. Spatial oblivious design (left 3 graphs) uses constant worse-case values comparing to spatial aware design (right 3).}
\label{fig:spatial_aware_design_velocity_visbility}
\end{figure*}

\textit{Space Traversal Speed:} Speed directly impacts the processing deadline. Higher speeds shorten the time available to dodge new obstacles, requiring shorter decision deadlines (Figure~\ref{fig:visibility_velocity_time_budget_relationship}).

\textit{Space Visibility:} Visibility (measured in meters) depends on how occluded a MAV's view is due to obstacles or weather conditions (i.e., blue sky vs. fog). Visibility impacts the processing deadline as the further a MAV can see, the more time it has to spot and plan around obstacles. 
For example, Figure~\ref{fig:visibility_velocity_time_budget_relationship}'s top curve has a higher deadline than the second to top curve, regardless of velocity, as it has a higher visibility.

\textit{In short, space precision and volume determine the processing latency while space visibility and traversal speed impose the deadline.
The above four features, and thus latency and deadline, are traditionally determined statically at design time. Next, we show how determining them dynamically can improve performance.}

\subsection{The Need for Spatial Aware (compute-space) Co-Design}


Traditional cyber systems are designed in a spatial oblivious manner, ignoring space demands during the mission time, and instead making worst-case, static assumptions about the environment, to guarantee safety. 
For example, if a portion of a mission demands high precision navigation, a constant worst-case precision is enforced, leading to a high latency throughout. Similarly, if a mission requires a high velocity, a permanent short deadline is enforced. 
In contrast, a spatial aware design dynamically captures the environment's heterogeneity and adapts cyber resources accordingly.


Figures~\ref{fig:spatial_aware_design_res_volume} and~\ref{fig:spatial_aware_design_velocity_visbility} exemplify two scenarios in which a spatial aware runtime can improve overall mission performance. Each figure shows a top-down view of the environment, with the black squares signifying obstacle positions and colored squares signifying the MAV's traversed trajectory. The heatmap shows the space feature's value, e.g., Figure~\ref{fig:static_precision} shows precision. In each figure, the left three graphs show the baseline, spatial oblivious runtime, and the right three show the spatial aware runtime. Note that although trajectories are different, as the planner is stochastic, the trends are worth analyzing. Also, although these examples are simple to keep the discussion concise, a thorough analysis of complex environments is provided in Section~\ref{sec:results}.

\begin{figure}[t!] 
    \vspace{-3em}
    \centering
    \begin{subfigure}{.44\columnwidth}
        \centering
        \includegraphics[trim=3 0 64 0, clip, width=\linewidth]{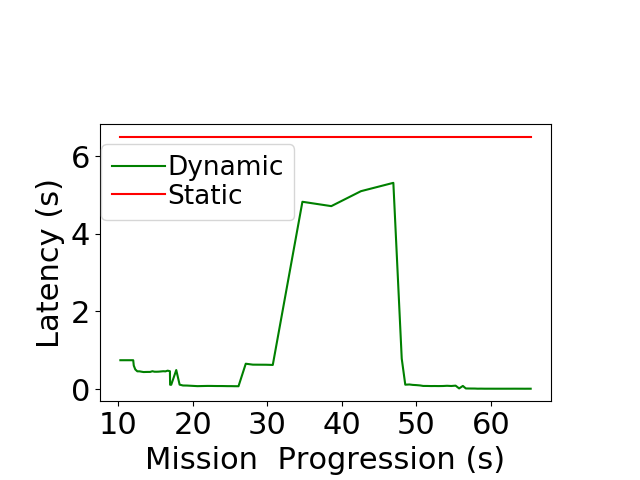}
       \caption{Latency. Lower is better.}
        \label{fig:response_time_slack}
    \end{subfigure}
    \hspace*{.2em}
    \begin{subfigure}{.44\columnwidth}
        \centering
        \includegraphics[trim=3 0 64 0, clip, width=\linewidth]{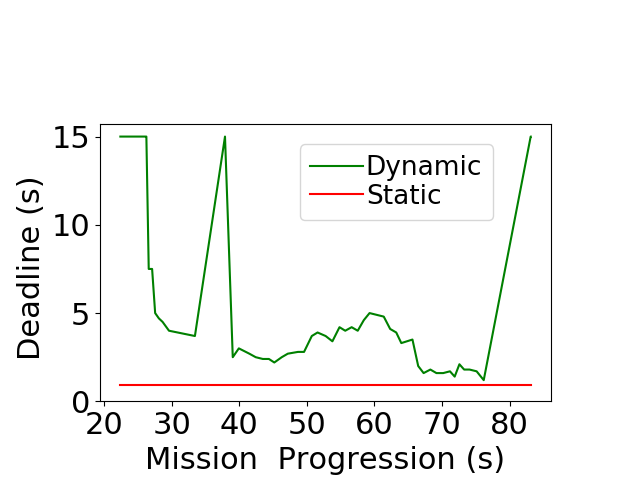}
       \caption{Deadline. Higher is better.}
        \label{fig:time_budget_slack}
    \end{subfigure}
\caption{Worst-case, static design vs dynamic (spatial-aware) design.}
\label{fig:res_quan}
\end{figure}

\textbf{High Precision Mission.} Figure~\ref{fig:spatial_aware_design_res_volume} emulates a high precision mission, e.g. a package delivery MAV flying through warehouse's tight aisles. The spatial oblivious design's constant worst-case precision (Figure~\ref{fig:static_precision}) and volume (Figure~\ref{fig:static_volume}) results in a high and fixed decision latency  (Figure~\ref{fig:static_res_vol_resp_time}). This high decision latency degrades the MAV's agility to dodge obstacles, lowering its safe velocity.

In contrast, the spatial aware design adjusts its precision (Figure~\ref{fig:dynamic_precision}) and volume (Figure~\ref{fig:dynamic_volume}) according to the space demands, varying its decision latency accordingly (Figure~\ref{fig:dynamic_res_vol_resp_time}). Precision is increased (voxel size decreased) as the MAV nears obstacles.  Similarly, a smaller volume is processed near obstacles as they momentarily occlude the MAV's field of view (FOV). 

Figure~\ref{fig:response_time_slack} shows the impact of such variations on the latency of the dynamic, spatial aware design (bottom curve) as compared to the static, spatial oblivious design (top curve). As we will show later, spatial aware design's lower latency improves safe velocity. Note that the spatial aware design's latency never reaches the spatial oblivious design's latency since there is never a demand for both high precision and high volume simultaneously, because near obstacles where high precision is demanded, the volume is reduced due to occlusions.


\textbf{High Velocity Mission.} Figure~\ref{fig:spatial_aware_design_velocity_visbility} emulates a mission with high velocity demands (e.g., search and rescue). A spatial oblivious design conservatively assumes high velocity (Figure~\ref{fig:static_velocity}) and low visibility (Figure~\ref{fig:static_visibility}), and hence a short deadline (Figure~\ref{fig:static_velocity_budget}). Short decision deadlines can not afford the high precision (long latency computations) required near obstacles, preventing the MAV from seeing the gaps, leaving the ringed area and finishing its mission.

In contrast, a spatial aware design considers velocity (Figure~\ref{fig:dynamic_velocity}), visibility (Figure \ref{fig:dynamic_visibility}) changes to adjust its deadline (Figure~\ref{fig:dynamic_velocity_budget}). Thus, it affords high precision computation and leaves the ringed area.

Figure~\ref{fig:time_budget_slack} shows the processing deadline's variations for the dynamic spatial aware design compared to the static worst-case assumptions of the spatial oblivious design. The extended deadline can then be used toward high-precision navigation when needed.

\section{RoboRun: A Spatial Aware Middleware}
\label{sec:roborun_design}

We present \textit{RoboRun}, a spatial aware middleware
that exploits the robot’s heterogeneous environment and compute subsystem synergy. It
continuously assesses the external world's and MAV's internal demands, adjusting the computational resources accordingly.
Here we discuss the MAV's components and then detail the RoboRun runtime. 

\subsection{Drone System Stack}
\begin{figure}[!t]
\centering
\includegraphics[trim=0 0 0 0, clip, width=\linewidth]{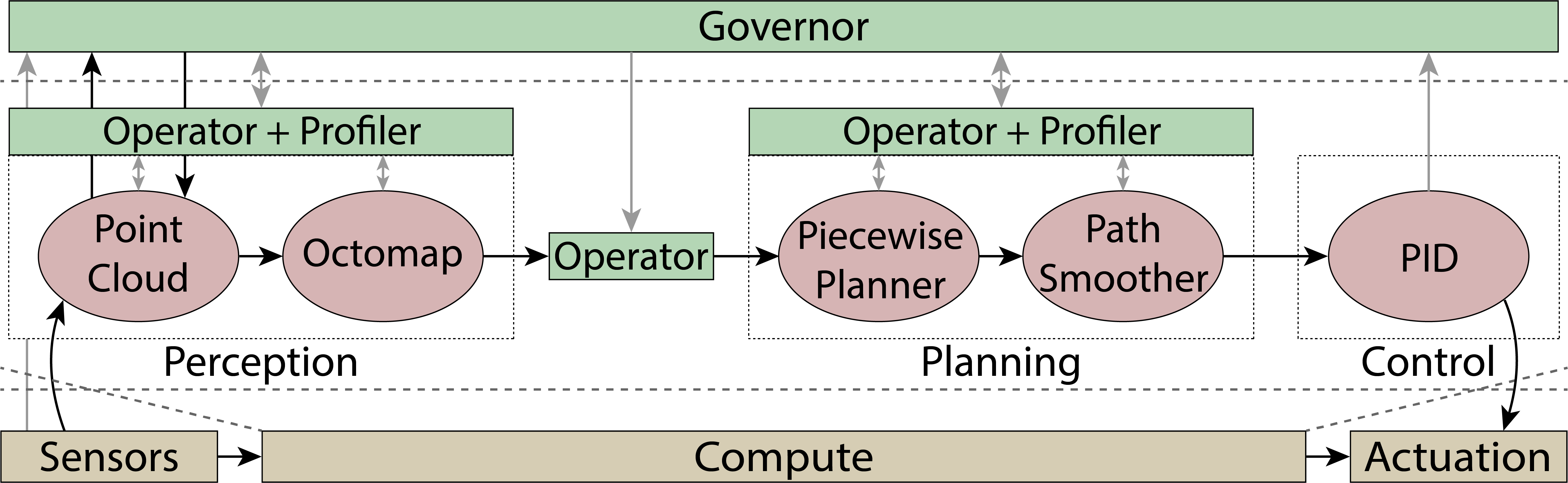}
\vspace{-5pt}
\caption{Drone's system stack: Hardware layer (brown), navigation pipeline (pink), runtime middleware where RoboRun sits (green). Black arrows show task flow, and gray ones show the dataflow in/out of runtime.}
\label{fig:PPC_pipeline}
\end{figure}

 \textbf{Hardware Layer:} Robots use sensors, actuators, and compute (Figure~\ref{fig:PPC_pipeline}, bottom, brown layer). 
Sensors capture the robot's states, compute makes navigational decisions, and actuators actualize them. We use a quadrotor MAV with 6 cameras, an IMU, and a GPS.

\textbf{Application Layer:} The navigation pipeline makes navigational decisions (e.g., direction, acceleration) using 3 fundamental robotics processing stages: \emph{Perception}, \emph{Planning}, and \emph{Control} (Figure~\ref{fig:PPC_pipeline}, middle, pink layer). \textit{Perception} interprets sensor information (e.g. camera images) to generate world models. We use the \textit{Point cloud} kernel to
extract obstacle positions by converting pixels to 3D coordinates.
The \textit{OctoMap} kernel then
accumulates these point clouds into a 3D map and encodes them in a tree data structure where each leaf is a \textit{voxel}. 

\textit{Planning}
generates a \textit{collision-free} path
using two kernels: \textit{piece-wise planning} and \textit{path smoothing}.
\textit{Piece-wise planning}
stochastically samples the map until a collision-free path to the destination is found.
We use the RRT* planner from the OMPL library due to its asymptotic optimality. 
We use Richter, et al.'s \textit{Path Smoothing} kernel to modify the piece-wise trajectory to incorporate the MAV's dynamic constraints such as maximum velocity~\cite{richter2016polynomial}.

\textit{Control} ensures that the MAV closely follows the generated trajectory while guaranteeing stability. 
We use standard PID control.

\textbf{Runtime Layer:} 
Runtime enforces the deadline and latency for each navigational decision. RoboRun sits in this layer and uses three components: a governor, profilers, and operators (Figure~\ref{fig:PPC_pipeline}, middle and top green blocks). The governor computes optimal time budgeting policies based on the MAV's internal and external states (e.g., velocity and obstacle density), which are monitored by profilers.
These policies are passed to the operators (gray dataflow arrows) for enforcement. We implemented RoboRun on top of the
Robot Operating System (ROS), which provides inter-process communication and common robotics libraries.
Next, we detail RoboRun's components.

\subsection{RoboRun Components: Operators}
Operators enforce the precision and volume policies. 
Due to space limitations, we only provide an overview of each. 

\textbf{Precision Operators:}
They enable accuracy-latency trade-offs by enforcing the precision of space processed. Environments with higher precision requirements incur greater computational latency.

Point cloud (perception) precision is enforced by controlling the sampling distance between points. We grid the space into cells, map the points onto the cells using their coordinates, and then reduce each cell to a single average point. 
OctoMap (perception) precision is enforced by controlling the step size of the raytracer.
Perception to planning precision is enforced by sub-sampling and pruning the tree structure of the encoded map, and planning precision is enforced by modifying the raytracer, similar to OctoMap.



\textbf{Volume Operators:}
They enable volume-latency trade-offs by enforcing the volume of space processed. Processing a higher volume leads to better world modeling but higher computational latency.

We deploy three volume operators.
The first operator (perception) controls the volume of space added to the map. We sort the space based on the distance to the MAV's trajectory, as closer points pose more threats if containing obstacles. Sorted points are integrated one by one until their resulting volume exceeds the desired threshold.
The second operator (perception to planning) controls the space volume communicated to the planner, limiting the planner's knowledge of the world.
Similarly, we sort the space based on the proximity to the MAV.  To do so, we prune the map, encoded in a tree, by selecting higher level trees (in the sorted order) until the threshold is reached.
The third operator (planning) controls the space volume explored by the planner. RRT* sorts the points/paths within the explored space and our volume monitor stops the search upon exceeding the threshold.
\subsection{RoboRun Components: Profilers}
To adjust precision and volume knobs, environmental information, e.g., gaps between obstacles, and internal drone states, e.g., velocity, are profiled from the sensors and navigation pipeline.
Profilers post-process each stage's data structures, e.g., point cloud array, tree map, and trajectory to extract space characteristics (Table~\ref{fig:profiling_table}).

\begin{table}[b]
\resizebox{.47\textwidth}{!}{%
\centering
\small
\begin{tabular}{|c|c|c|}
\hline
\textbf{Variable}            & \textbf{Profiled Pipeline Stage} & \textbf{Used For}       \\ \hline 
Gap between obstacles        & Point cloud               & Precision             \\ \hline
\begin{tabular}[c]{@{}c@{}}Closest obstacle, \\ closest unknown\end{tabular} & \begin{tabular}[c]{@{}c@{}}Point cloud,  \\ OctoMap, Smoother\end{tabular}      & \begin{tabular}[c]{@{}c@{}}Precision, Volume, \\ Deadline\end{tabular}  \\ \hline
 Sensor, map volume        & Point cloud, OctoMap               & Volume                  \\ \hline
 Velocity, position             & Sensors                   & Deadline            \\ \hline
Trajectory             & Smoother                & Deadline            \\ \hline
\end{tabular}}
\vspace{1pt}
\caption{The variables collected by profilers.}
\label{fig:profiling_table}
\end{table}

\subsection{RoboRun Components: Governor}
\label{subsec:governor}
The governor allocates the time budget (deadline) for the end-to-end navigation pipeline and determines the correct precision and volume settings per stage to satisfy this budget and space demands. Next, we discuss the governor's time budgeting algorithm and solver.

\subsubsection{Time Budgeting}
The \textit{time budget} (deadline) is the maximum time the MAV can spend processing a sampled input while ensuring a safe flight. As shown in Equation~\eqref{eq:motion_model}, it depends on $v$, the space traversal velocity, $d$, the space visibility, and $d_{stop}(v)$, the distance the MAV needs to stop when traveling at velocity $v$~\cite{liu2016high}. The numerator specifies the minimum safe distance between an obstacle and the MAV given its velocity. The division by the denominator sets an upper bound on the time budget. 
$d_{stop}(v)$ is modeled by flying the drone with various velocities in simulation and measuring the stopping distance. Equation \ref{eq:d_stop} shows our model with 2\% MSE. 

\begin{equation}
  \vspace{-10pt}
    \label{eq:motion_model}
    \text{Time budget} = \frac{(d - d_{stop}(v))}{v}
\end{equation}

\begin{algorithm}[!t]
\begin{algorithmic}[1]
\caption{Time Budgeting Algorithm}
\label{fig:budgetting_algorithm}
\State $b_g \gets 0, b_r \gets \text{Equation}~\ref{eq:motion_model} \text{ at } W_0$
\For{$i = 1 \dots |W|$}
    \State $b_r \gets b_r - flightTime(i,i-1)$
    \State $b_l \gets \text{Equation}~\ref{eq:motion_model} \text{ at } W_i$
    \State $b_r \gets min(b_r,b_l)$
    \If{$b_r \leq 0$} \textbf{break}
    \EndIf
    \State $b_g \gets b_g + flightTime(i,i-1)$
\EndFor
\State \Return $b_g$
\end{algorithmic}
\end{algorithm}

\begin{equation}
   \vspace{-3pt}
    \label{eq:d_stop}
    d_{stop(v)}= -0.055*v^2 - 0.36*(v)+ 0.20
\end{equation}

Equation~\eqref{eq:motion_model} assumes a fixed velocity and visibility for the duration of the time budget. This can result in overly optimistic budgets as these variables can change. To improve this model, we derive an algorithm to take into account the instantaneous velocity and visibility, and the planned velocity and visibility for upcoming waypoints ($W$). Our algorithm computes the global\_budget ($b_g$) as a running sum. This ensures that at each waypoint, the duration of the flight from the previous waypoint and the local\_budget at that waypoint ($b_l$ computed using Equation~\eqref{eq:motion_model}) are considered.
With this, environment variations at all waypoints are accounted for, ensuring a safe flight. See Algorithm~\ref{fig:budgetting_algorithm} for more details.


\subsubsection{Solver}
It uses the profiled values and sets each stage precision/volume knobs to fit the time budget and ensure space demands.

We formulate our problem as a constrained optimization problem (Equation~\eqref{eq:opt_problem}). 
The objective is to minimize the difference between $\delta_d$, the desired end-to-end latency acquired from the time budgeter, and $\sum_i \delta_{i}(p_i,v_i)$, the sum of the latency across all application layer stages, $i$, as a function of their precision and volume $p_i, v_i$. Within the application layer, we have 3 stages, i.e., perception ($i=0$), perception to planning ($i=1$), and planning ($i=2$).

The precision and volume of each stage are subject to constraints. The perception volume is limited by the perception to planning volume, which is itself limited by $v_{\text{sensor}}$ and $v_{\text{map}}$, the max volume ingested by the sensors and map, respectively. The perception precision is limited by $g_{\text{avg}}$ and $g_{\text{min}}$, the average and minimum gap size of the observed volume of space, respectively, and $d_{\text{obs}}$, the distance of MAV to the nearest obstacle. Our framework also requires the precision for the perception to planning and planning to be equivalent.
Due to the OctoMap framework's constraints, the precision used must also be a power of two multiple of the minimum voxel size, $vox_{\text{min}}$. 
\vspace{-3pt}
\begin{align} \label{eq:opt_problem}
    & \min_{p,v} \left(\delta_d - \sum_i \delta_i(p_i,v_i)\right)^2 \\
    & g_{\text{min}} \leq p_0 \leq \min(p_1, g_{\text{avg}}, d_{\text{obs}}) \nonumber \\
    & v_0 \leq v_1 \leq \min(v_{\text{sensor}}, v_{\textit{map}}) \nonumber\\
    & p_i \in \{ vox_{\min} 2^n : 0 \leq n \leq d-1 \} \nonumber
\end{align}





To find each stage's latency as a function of precision and volume, $\delta_i(p_i, v_i)$, we profiled a representative set of precision-volume combinations. 
We then fit a polynomial model to this data with $<$8\% average MSE. Our model is shown in Equation~\ref{eq:model}, with coefficient vector $\textbf{q}_i \in \mathbb{R}^4$ (where $\hat{p}_i = 1 / p_i$. This change of variables improves the numerical conditioning of the optimization problem).
\begin{align}
    \delta_i(p_i,v_i) = (q_{i,0} \hat{p}_i^3 + q_{i,1} \hat{p}_i^2 + q_{i,2} \hat{p}_i) (q_{i,3} v_i) 
    \label{eq:model}
\end{align}

\section{Evaluation Setup and Methodology}
\label{sec:evaluation}
In this section, we detail our evaluation setup.


\textbf{HIL Simulation:} We simulate the MAV's environment using the Unreal game engine, a physics and rendering
engine, and use AirSim, an Unreal Plugin to simulate the MAV's actuation/sensors.
We use a ``hardware-in-the-loop'' (HIL) methodology where our environment simulation and workload run on different machines improving simulation fidelity. The former runs on an Intel Core i9 and an NVIDIA 2080 Ti, and the latter on four Core i9 cores.


\textbf{Missions:}
To capture the characteristics of most missions, we emulate two fundamental MAV scenarios: \textit{package delivery}, where products are transferred from one warehouse to another, and \textit{search and rescue}, where medical equipment is transferred from a hospital to patients stranded in a disaster zone.

\textbf{Environment Generation:}
To emulate real-life missions, we developed an environment generator to systematically vary space difficulty/heterogeneity.
Our generator adjusts environment difficulty with hyperparameters that change the number of congestion clusters, obstacle density, and spread.
Obstacle density determines the ratio of occupied cells around a grid cell, and the obstacle spread impacts the space area around a cluster center where obstacles are spawned.
A Gaussian distribution uses these parameters to generate 27 different environments used for the result section.

\begin{table}[]
\centering
\small
\begin{tabular}{|c|c|c|}
\hline
\textbf{Variable}                                                           & \textbf{Static} & \textbf{Dynamic} \\ \hline
Point cloud precision ($m$)                                                 & 0.3                   & \begin{tabular}[c]{@{}c@{}}{[}0.3 \ldots 9.6{]}\end{tabular}         \\ \hline
\begin{tabular}[c]{@{}c@{}}OctoMap to planner precision ($m$)\end{tabular} & 0.3                   & \begin{tabular}[c]{@{}c@{}}{[}0.3 \ldots 9.6{]}\end{tabular}         \\ \hline
OctoMap volume ($m^3$)                                                         & 46000                 & {[}0 \ldots 60000{]}         \\ \hline
\begin{tabular}[c]{@{}c@{}}OctoMap to planner volume ($m^3$)\end{tabular}    & 150000                & {[}0 \ldots 1000000{]}       \\ \hline
Planner volume ($m^3$)                                                         & 150000                & {[}0 \ldots 1000000{]}       \\ \hline
\end{tabular}
\vspace{1pt}
\caption{Knob value(s) used subject to constraints in Eq. \eqref{eq:opt_problem}.}
\label{fig:knob_setting_values}
\end{table}
\textbf{Baseline and Knob Values:}
We use the state-of-the-art navigation pipeline provided in MAVBench~\cite{boroujerdian2018mavbench} as the static, spatial oblivious baseline. For the baseline, knobs are set such that the mission can be successfully executed, i.e., with a precision to allow navigating narrow real-world aisles~\cite{flytware, flytbase, flytware_article}, and with volumes to allow MAV to collect all 6 camera data and generate maps matching an average warehouse size~\cite{warehouse_size}. For RoboRun's spatial aware design, the knob values vary dynamically, exploiting the spatial heterogeneity while ensuring safety.
In both cases, the maximum velocity is chosen experimentally such that at least 80\% of flights are collision-free. Knob values/ranges are shown in Table~\ref{fig:knob_setting_values}.
\section{Results} \label{sec:results}

We compare RoboRun against the state-of-the-art static/spatial oblivious navigation pipeline~\cite{boroujerdian2018mavbench}. Here, we present our overall results, an evaluation of RoboRun's robustness across environment difficulty levels, and finally, a thorough analysis of a representative mission. 




\subsection{Overall Results}

\begin{figure}[h]
\vspace{-.55em}
\hspace{-2em}
\centering
\includegraphics[width=0.99\columnwidth]{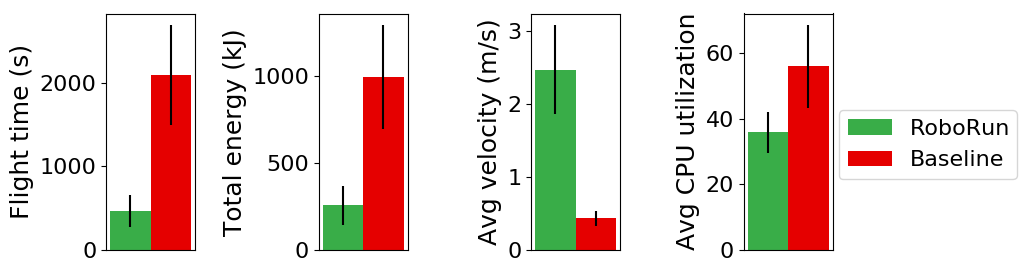}
\caption{Mission level metrics for two designs.}
\label{fig:mission_metrics}
\end{figure}

We presents results averaged over 27 environments (Figure~\ref{fig:mission_metrics}).

\textbf{Flight Velocity.} By adjusting precision and volume knobs continuously, RoboRun improves velocity by 5X (from 0.4~m/s to 2.5~m/s). Concretely, by lowering the precision and volume when possible, RoboRun reduces each decision's latency, improving the drone's reaction time, and enabling a higher velocity yet collision-free flights.


\textbf{Flight Time.} RoboRun's higher velocity enables a faster package delivery or search and rescue mission. The aforementioned 5X improvement in velocity allows the drone to traverse the space faster to arrive at its destination. RoboRun improves the overall mission time by as much as 4.5X (from 2093~(s) to 465~(s)). 


\textbf{Flight Energy.} RoboRun also improves energy consumption by 4X (from 1000~kJ to 257~kJ). This improvement occurs because flight energy is highly correlated with flight time, as propellers consume large amounts of energy even when hovering, i.e., zero velocity \cite{boroujerdian2018mavbench}. Therefore, lowering flight time reduces energy usage. Note that since compute consumes less than 0.05\% of the overall MAV's energy \cite{boroujerdian2018mavbench}, reducing compute energy consumption does not directly impact the overall mission's energy. \textit{Instead, compute's impact is through increasing velocity, reducing flight time, and hence, flight energy.}

\textbf{System Utilization.} RoboRun reduces CPU-utilization by 36\% on average per decision by lowering the computational load when possible. This frees up CPU resources for higher-level cognitive tasks, e.g., semantic labeling, and gesture/action detection. Since navigation is a primitive task, lowering its pressure on the CPU is imperative.

\subsection{Robustness Across Environment Heterogeneity}
To show robustness and generality, we evaluate Roborun within environments of different difficulty and heterogeneity levels. 

\begin{figure}[t!]
\vspace{-.75em}
\centering
\hspace{-2em}
    \begin{subfigure}{0.53\columnwidth}
        \resizebox{\columnwidth}{!}{
            \begin{tabular}{|c|c|}
            \hline
            \textbf{Environ. Knobs} & \textbf{Dynamic Values} \\ \hline
            Obstacle Density            & {[}0.3, 0.45, 0.6{]}    \\ \hline
            Obstacle Spread   (m)        & {[}40, 80, 120{]}       \\ \hline
            Goal Distance (m)             & {[}600, 900, 1200{]}    \\ \hline
            \end{tabular}
        }
        \caption{Evaluation scenarios.}
        \label{fig:eval_scenarios}
    \end{subfigure}
    \vspace{-0.5em}
    \begin{subfigure}{0.51\columnwidth}
        \vspace{-1em}
        \centering
        \includegraphics[trim=0 0 0 0, clip, width=\linewidth]{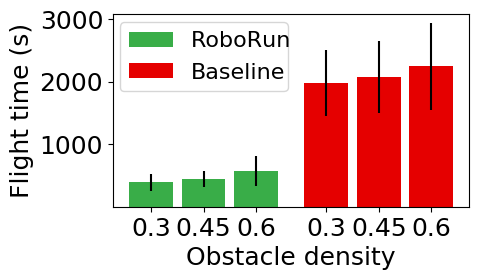}
        \vspace{-2em}
        \caption{Sensitivity to density.}
        \label{fig:env_sensitivity_density}
    \end{subfigure}
    \begin{subfigure}{0.51\columnwidth}
        \centering
        \hspace{-1.5em}
        \includegraphics[trim=0 0 0 0, clip, width=\linewidth]{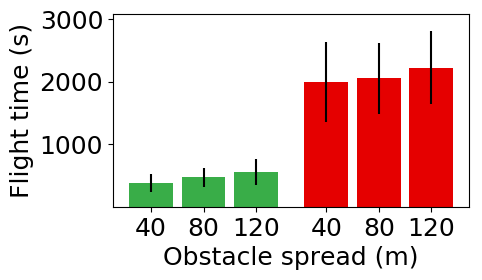}
        \vspace{-0.9em}
        \caption{Sensitivity to spread.}
    \label{fig:env_sensitivity_spread}
    \end{subfigure}
    \begin{subfigure}{0.47\columnwidth}
        \vspace{0.3em}
        \centering
        \includegraphics[trim=-50 0 0 0, clip, width=\linewidth]{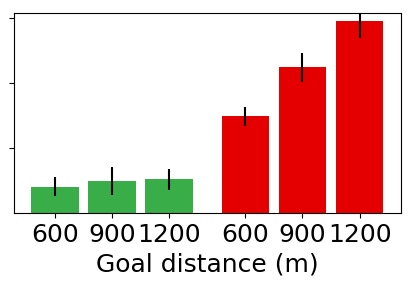}
        \vspace{-2em}
        \caption{Sensitivity to goal distance.}
        \label{fig:env_sensitivity_gd}
    \end{subfigure}
    \caption{Sensitivity analysis across different difficulty knobs.}
    \label{fig:env_sensitivity}
\end{figure}

Our environment generator controls space difficulty using obstacle density, spread, and goal distance (3 values for each, Figure~\ref{fig:eval_scenarios}) to generate 27 different environments. Each randomly generated environment contains two congested (A and C) zones and one non-congested (B) zone. Congested zones are located at the beginning and end of the mission to emulate warehouse-building or hospital-building combinations. These zones are heterogeneous, with a gradual reduction of congestion outward from their center. In contrast, zone B is homogeneous and bigger, representing a longer distance traveled, either in open skies or over a city. Figure ~\ref{figs:congestion_map} shows a sample mission.




\textbf{Obstacle Density:}
The peak density of obstacles impacts the space difficulty by affecting the obstacle count and their relative distance from each other. A higher obstacle density negatively impacts mission time and energy (Figure~\ref{fig:env_sensitivity_density}) by resulting in more closely placed obstacles, demanding a higher precision, higher latency computation, and, hence, a lower safe velocity. RoboRun and the baseline experience a worst-case 1.5X and 1.1X increase in flight time, respectively, across the highest and lowest density environments. RoboRun has a higher sensitivity to this knob as it exploits easier environments whereas the poorer performance of the baseline is less impacted.
Energy closely follows flight time and is omitted due to space.

\textbf{Obstacle Spread:} It determines the radius within which obstacles are spread. RoboRun and the baseline experience an increase in flight time by as much as 1.4X and 1.1X respectively, across the highest and lowest spread environments (Figure~\ref{fig:env_sensitivity_spread}). 
The spread has a smaller impact comparing to density as it impacts the area over which obstacles are distributed as opposed to their density. 

\textbf{Goal Distance:}
As goal distance increases, so does mission time. Goal distance increases flight time for RoboRun and the baseline by as much and 1.3X and 2X, respectively (Figure~\ref{fig:env_sensitivity_gd}). RoboRun minimizes the impact of the increased goal distance by swiftly flying in non-congested areas. However, the baseline is heavily impacted by the goal distance due to its conservative, low max velocity, making long-distance flights, often necessary for package delivery and search and rescue, infeasible as longer flight times expend the battery.

\subsection{Representative Mission Analysis}

\begin{figure}[t]
\centering
\includegraphics[width=0.99\columnwidth]{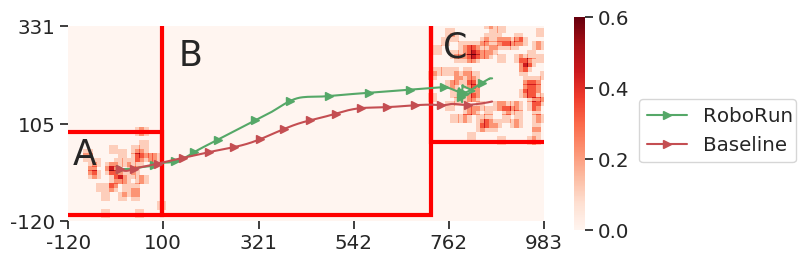}
\caption{\small Mission example. X, Y axis are map coordinates. Heat map shows each point's congestion level. Arrows are the trajectories travelled.
}
\label{figs:congestion_map}
\end{figure}

\begin{figure}[b!]
\hspace*{-0.5em}
\centering
    \begin{subfigure}{1.05\columnwidth}
        \centering
        \includegraphics[trim=0 0 0 0, clip, width=\linewidth]{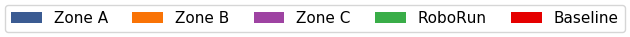}
    \end{subfigure}
   
\hspace*{-0.9em}
    \scalebox{1.1}{ 
        \begin{subfigure}{0.27\columnwidth}
            \includegraphics[valign=t, trim=5 7 0 8, clip, width=\linewidth]{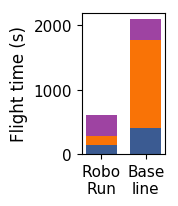}
            \caption{Flight time.}
            \label{fig:sample_flight_time_energy}
        \end{subfigure}
        \begin{subfigure}{0.33\columnwidth}
            \includegraphics[valign=t, trim=0 0 0 10, clip, width=\linewidth]{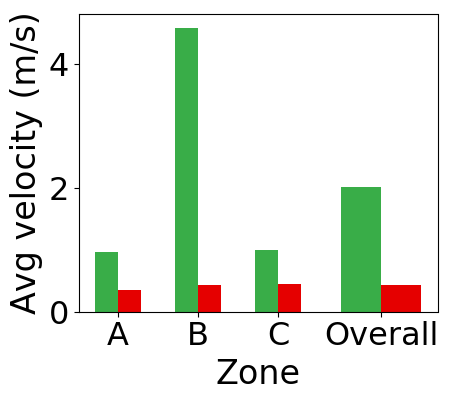}
            \caption{Velocity.}
            \label{fig:sample_flight_avg_velocity}
        \end{subfigure}
        \begin{subfigure}{0.37\columnwidth}
            \includegraphics[valign=t, trim=0 8 0 8, clip, width=\linewidth]{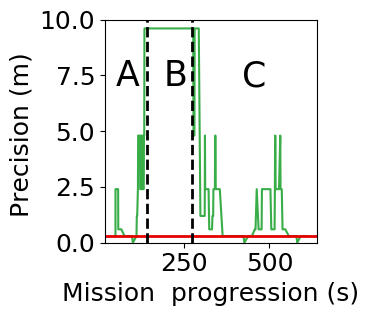}
            \caption{Precision over time.}
            \label{fig:precision_dynamic_time_prog}
        \end{subfigure}
    }
    \label{fig:sample_flight_metrics}
    \caption{Time, velocity, and precision for a representative mission.}
\end{figure}

To better understand how RoboRun arrives at the previously discussed improvements, we analyze a representative mission in detail. Our mission takes place in an environment with the mid-range difficulty level (as shown in Figure~\ref{figs:congestion_map}) 
to avoid skewing the result toward either extreme of environment difficulty. Although the trajectories across two runs (even of the same design) vary due to the planner's stochastic nature, the trends are worth analyzing.  


\textbf{Mission Metrics:}
RoboRun improves mission time and energy usage by around 3.5X and 3X respectively (Figure~\ref{fig:sample_flight_time_energy}). Note that RoboRun spends a shorter amount of time in zone B compared to the baseline as it can take advantage of this zone's low congestion.

\begin{figure*}[t]
\vspace*{-1.7em}
\centering
    \begin{subfigure}{0.7\textwidth}
            \centering
            \includegraphics[trim=0 0 0 0, clip, width=\linewidth]{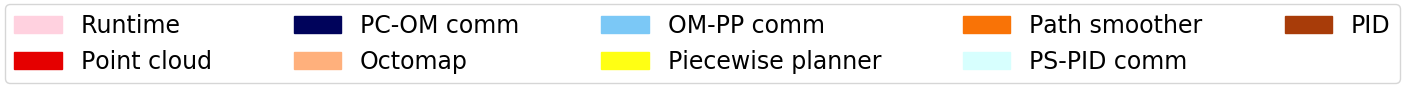}
            \includegraphics[trim=0 0 0 10, clip, width=\linewidth]{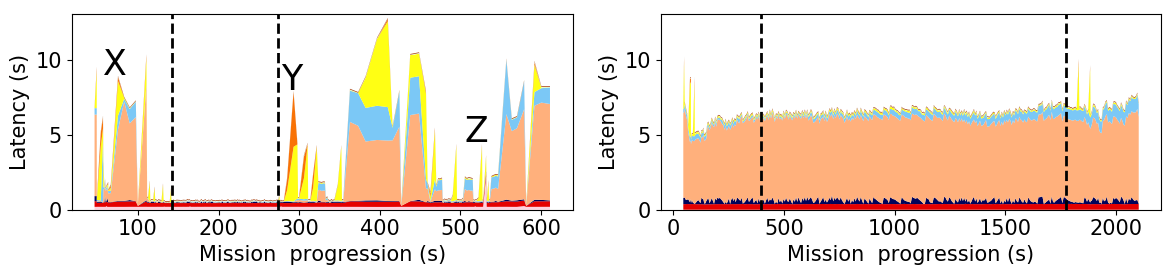}
            \caption{RoboRun (left) vs baseline (right) latency variations/breakdown through out the mission. Points of interest X, Y, and Z discussed below in the text. Dashes delimit zones A, B, C from left to right.}
            \label{fig:time_breakdown_dynamic_static}
    \vspace{-3pt}
    \end{subfigure}
    \begin{subfigure}{0.28\textwidth}
        \centering
        \includegraphics[trim=-75 0 0 -20, clip, width=0.93\linewidth]{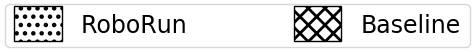}
        \includegraphics[trim=0 8 0 8, clip, width=\linewidth]{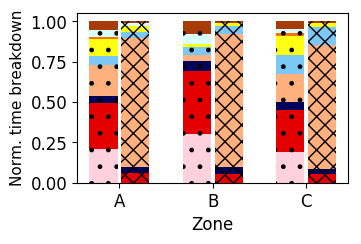}
        \caption{Normalized latency breakdown. Color code follows~\ref{fig:time_breakdown_dynamic_static}.}
        \label{fig:computation_hotspot}
     \vspace{-1pt}
    \end{subfigure}
    \caption{Latency breakdown for a representative mission. OM denotes OctoMap, PS denotes Path Smoother, and comm denotes communication.}
    \label{fig:ee_latency_breakdown_hotspot}
\end{figure*}

Reducing flight time requires increasing velocity. RoboRun improves velocity by 4.6X (0.43~m/s to 2.01~m/s) by adapting the space precision and volume to the zone it is traveling in (Figure~\ref{fig:sample_flight_avg_velocity}). 
Zone B's low congestion relaxes precision demands, reducing latency, and enabling higher velocity (4.6X) comparing to other zones. In contrast, the baseline maintains a constant lower velocity, ignoring space demands. Note that even within the congested zones (A/C), RoboRun reaches a higher velocity by exploiting intra zone heterogeneity.

\textbf{Precision and Volume Changes:}
RoboRun captures the heterogeneous precision/volume demands of different zones and uses them toward higher velocity.
Figure~\ref{fig:precision_dynamic_time_prog} shows the precision (Y-axis) changes at the decision level over time (X-axis) for both designs. Zones are delimited using dashed lines with zones A, B, C mapped to left, middle, and right sections.
RoboRun captures the greater heterogeneity of zones A/C with a significant variation in precision. Zone B sees no variation reflecting its homogeneity. In contrast, the baseline enforces the same worst-case precision throughout, ignoring space changes. RoboRun's highest precision matches those of the baseline, indicating the need for this worst-case precision. 
Volume variations have similar trends and are omitted due to space limitations.

\textbf{End-to-End Latency:}
The compounding impact of precision and volume varies the end-to-end latency of both designs. Figure ~\ref{fig:time_breakdown_dynamic_static} shows the end-to-end latency broken down by computation (shades of red) and communication (shades of blue) stages.
RoboRun (Figure~\ref{fig:time_breakdown_dynamic_static}, left graph) maintains an end-to-end latency less than the baseline's (Figure~\ref{fig:time_breakdown_dynamic_static}, right graph) throughout most of the mission with a median latency reduction of 11X. However, it incurs some high latency outliers near obstacles, as it aggressively reduces precision and consumes volume causing rare slowdowns of up to 1.2X. 

RoboRun also better matches the environment with almost no variation in the homogeneous zone (0.15 seconds for zone B) and high latency variations in the heterogeneous zones (10 and 12.5 seconds for zones A and C, respectively). In contrast, the baseline maintains a smaller change across all zones (max of 5.4 seconds). Note that the baseline cannot use its variations toward velocity improvement due to its fixed statically determined deadline.

 Baseline has a largely uniform computation/communication breakdown in comparison with RoboRun. For RoboRun, spikes can be categorized into high-precision/low-volume (e.g. point X at about 60 s), low-precision/high-volume (e.g. point Y at about 290 s), and mid-precision/mid-volume (e.g. point Z) moments. The first cluster (perception heavy) occurs in high congestion/high goal visibility/accessibility areas. The second (planning heavy) occurs in low congestion/low goal visibility, and the third somewhere in between. 

Both designs pay a fixed 210 (ms) point cloud latency, and RoboRun pays an extra 50 (ms) runtime latency. 

\begin{figure}[t]
\centering
\end{figure}

Figure ~\ref{fig:computation_hotspot} shows what portion of the end-to-end latency each stage consumes. For RoboRun the bottleneck shifts depending on the precision and the volume required. Zone B's low congestion shifts the bottleneck toward runtime and point cloud since big voxels reduce OctoMap's processing time, and in general, no re-planning is needed. However, congested areas within zones A and C distribute the computation more evenly across all stages. In contrast, the baseline (hashed bars) always distributes the computation the same way, pressuring OctoMap. Note RoboRun's more balanced pipeline in high congestion areas (i.e., A/C) when the system is under high pressure.

\section{Conclusion}
\label{sec:conclusion}
The limited onboard energy capacity of autonomous mobile robots requires efficient compute designs.
We propose \textit{spatial aware computing}, a compute-space co-design framework to exploit the synergy between a robot’s heterogeneous environment and compute subsystem. 
We identify 4 key features of space heterogeneity and present a spatial aware middleware (Roborun) to exploit them. Roborun increases the decision rate and velocity, and improving mission time and energy by 4.5X and 4X, respectively. 
It also improves CPU-utilization by 36\%, freeing CPU resources for higher-level cognitive tasks. 

\bibliographystyle{plain}


\end{document}


\title{Appendix}


\date{}
\maketitle

\thispagestyle{empty}

\section{Additional Evaluation Methodology Details}
Here we provide further details about our evaluation methodology and experimental setup. First, we discuss our environment generation details, and then we present various knobs used in both RoboRun and our baseline. 
\subsection{Environment Generation}

\begin{figure}[b!]
    \centering
    \includegraphics[width=0.75\columnwidth]{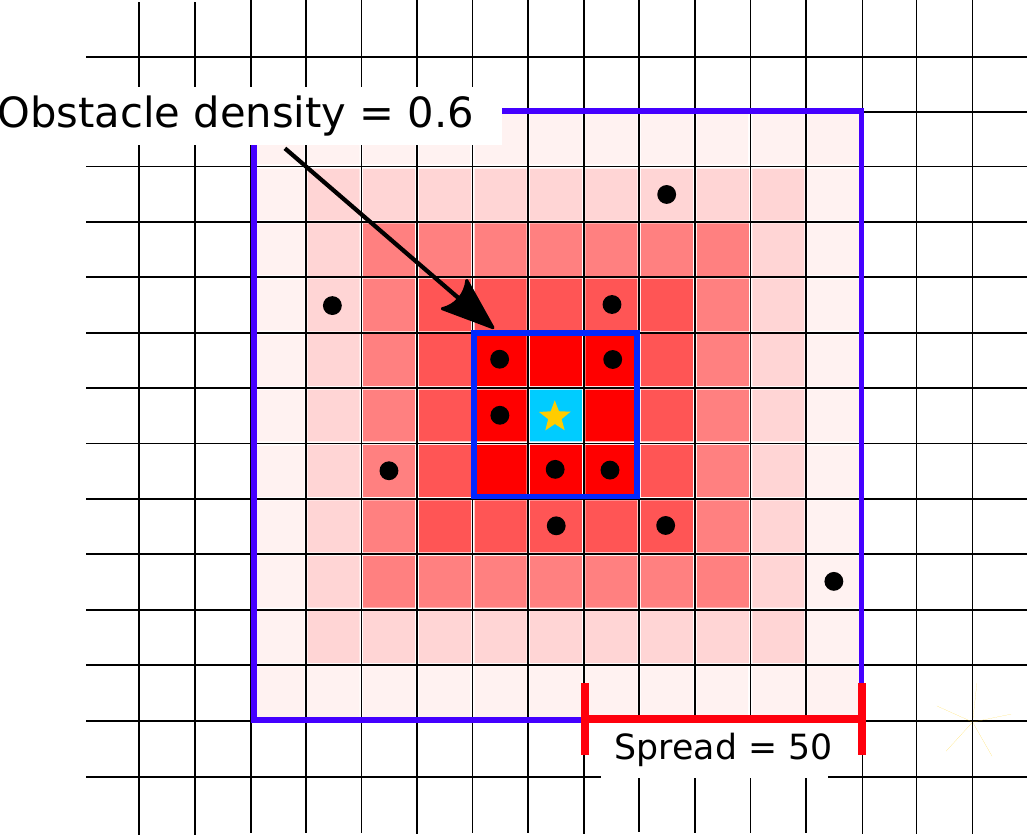}
    \caption{Obstacle density and spread around a centroid. Black dots indicate cylindrical pillar obstacles. The darker the shade of red, the higher the obstacle density.}
    \label{fig:env_gen_diagram}
\end{figure}

We developed an environment generator that systematically varies space difficulty/heterogeneity. This allows us to emulate various real-life missions accurately and further evaluate RoboRun. We believe that as the robotics system/OS community grows, generating these datasets in a principled way is imperative.

The environment generator is built on top of the Unreal game engine~\cite{GameEngi70:online}, and Air Learning Extensions~\cite{krishnan2019air}. It accepts a set of knobs, and we detail each here. 

\begin{figure}[h!]
    \centering
    \includegraphics[width=0.8\columnwidth]{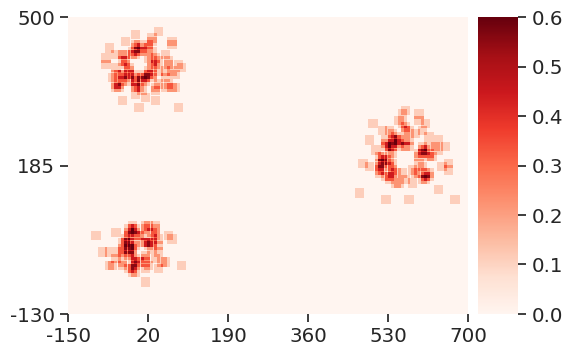}
    \caption{Environment generated with multiple centroids.}
    \label{fig:multiple_centroids}
\end{figure}

\textbf{Centroids}. Obstacles are spawned around centroids.
The blue cell (marked with a star) in Figure~\ref{fig:env_gen_diagram} denotes the centroid. Figure~\ref{fig:multiple_centroids} shows an environment with multiple centroids.
This knob impacts environment heterogeneity. 

\textbf{Centroid distance}. This determines the distance between the two landing positions of the drone. This knob impacts mission difficulty.

\textbf{Grid size}. We grid the arena into squares of length \texttt{GridSize}. Figure \ref{fig:env_gen_diagram} shows an environment with grid size of 10 m. This knob impacts the precision of the environment generator and the size of each obstacle impacting environment difficulty.

\textbf{Gap size}. This determines the amount of free space between two obstacles (cylindrical pillars in our case). Figure~\ref{fig:env_gen_diagram} shows an environment with gap size of 6 m. This impacts environment difficulty. 

\textbf{Obstacle density}. 
This determines the ratio of free and occupied grid cells surrounding a point in space.
Intuitively, points with a higher density are more difficult to navigate through as they demand a higher precision. 
Figure~\ref{fig:env_gen_diagram} shows obstacle density of 60\% around the centroid.
This knob impacts environment difficulty. 

\textbf{Spread}. This determines the size of the area in which we spawn obstacles. Areas with higher spread contain more obstacles rendering them more challenging to navigate within. Figure~\ref{fig:env_gen_diagram} shows an environment with a spread of 50 meters around the centroid. This knob impacts environment difficulty and heterogeneity. 



\textbf{Distribution Model.} This is the probability distribution with which we spawn obstacles around centroids. This impacts environment heterogeneity.

\textbf{Putting All Together.} In RoboRun, we use a Gaussian distribution to spawn obstacles around the start and the goal positions with mean of \texttt{ObstacleDensity}, progressively radiating out with decreasing probability to the distance of \texttt{Spread}. 

\begin{figure}[t!]
\centering
    \begin{subfigure}{0.3\columnwidth}
        \includegraphics[width=\linewidth]{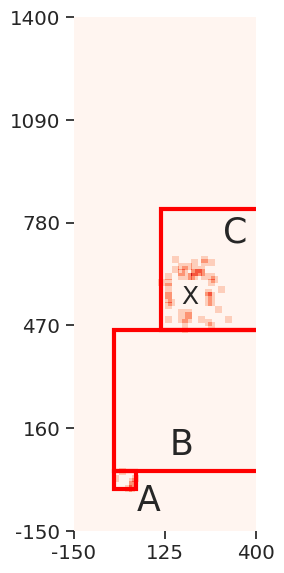}
        \label{fig:env_gen_easy}
    \end{subfigure}
    \begin{subfigure}{0.3\columnwidth}
        \includegraphics[width=\linewidth]{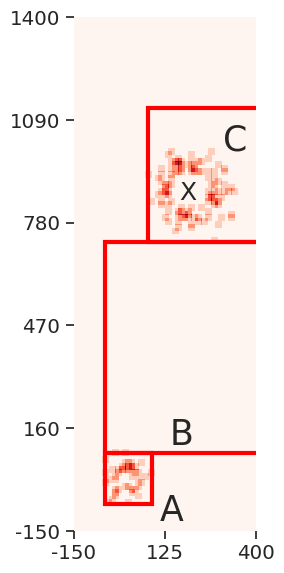}
        \label{fig:env_gen_medium}
    \end{subfigure}
    \begin{subfigure}{0.35\columnwidth}
        \includegraphics[width=\linewidth]{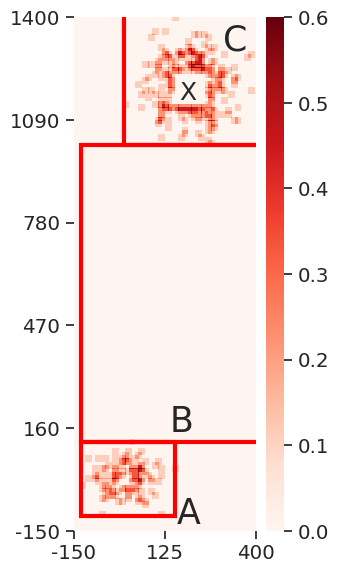}
        \label{fig:env_gen_hard}
    \end{subfigure}
    \vspace{-1em}
    \caption{Easiest, medium, and hardest environments. X~marks the drone's goal location. It starts in the center of Zone A.}
    \label{fig:env_gen_examples}
\end{figure}

\begin{figure}[h!]
    \centering
    \begin{subfigure}{0.8\columnwidth}
        \includegraphics[width=\linewidth]{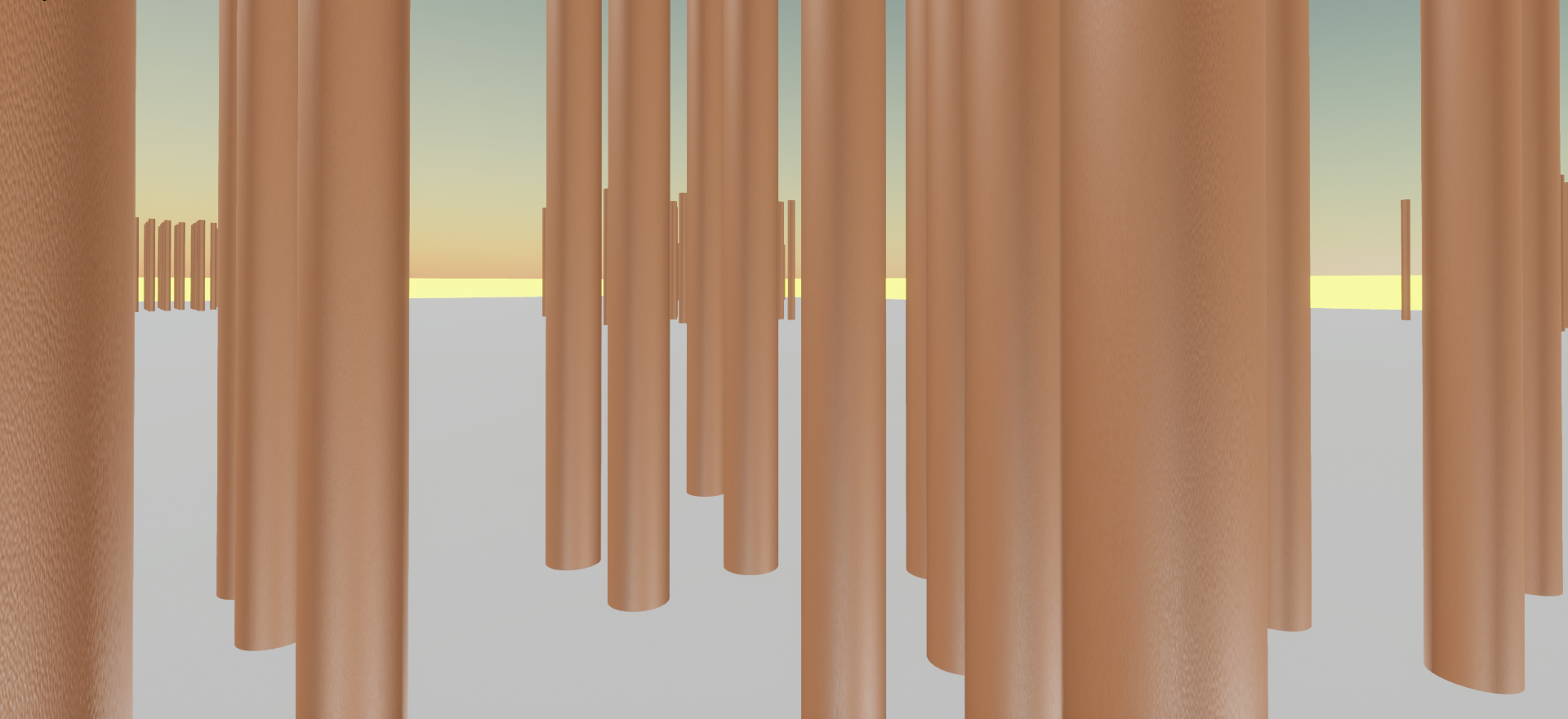}
        \caption{Drone's point-of-view.}
        \label{fig:drone_fpv}
    \end{subfigure}
    \begin{subfigure}{0.8\columnwidth}
        \includegraphics[width=\linewidth]{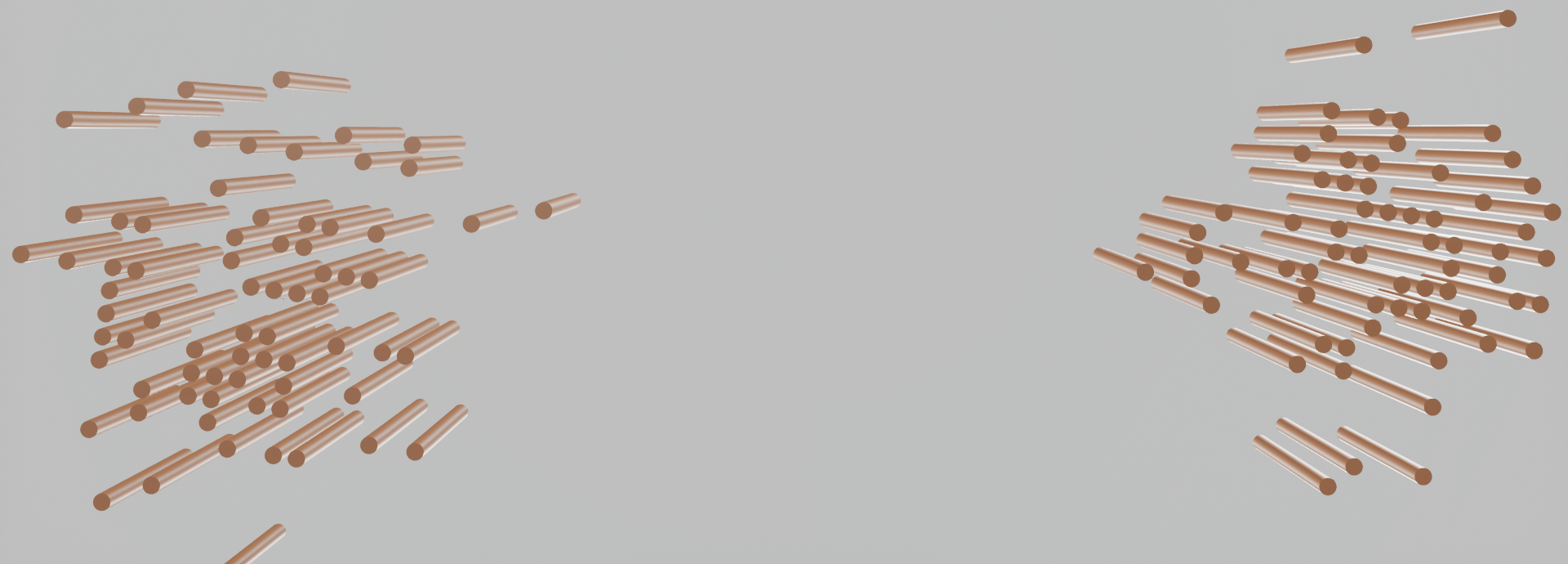}
        \caption{Aerial view.}
        \label{fig:aerial_view}
    \end{subfigure}
    \caption{Simulation environment.}
    \label{fig:simulation_env}
\end{figure}

Figure \ref{fig:env_gen_examples} shows three example environments generated with the easiest, medium, and hardest settings. Figure \ref{fig:simulation_env} shows a snapshot of the simulation environment.


\subsection{Knob Values Used for RoboRun and Baseline}
Both RoboRun and the spatial oblivious baseline are configured with a set of knobs such that the mission can be successfully executed. These values are provided in Table~\ref{fig:knob_setting_values}. 

For both designs the highest precision (i.e. smallest voxel size) is 0.3m, to allow the MAV to successfully pick up packages/medical equipment from shelves in narrow real-world aisles~\cite{flytware, flytbase, flytware_article}.\footnote{$\frac{1.8 (standard\;narrow\;aisle\;width) - 1.2 (drone\;radius)}{2} =0.3$ Note the division by 2 is because each side of the aisle can be bloated by a voxel size.} Note that although even higher precisions (i.e., lower voxels values that 0.3) can be desirable where close interactions with humans/machines and objects are necessary, 0.3 is the highest precision with which the baseline can successfully navigate. Voxel sizes of lower values impose velocities lower than 0.05 m/s, rendering the baseline ineffective to compare against. 
RoboRun additionally uses an upper bound of 9.6m, as values above this threshold provide no improvement to latency. 

Space volume for OctoMap under the baseline is 46000$m^{3}$, equal to the volume of space collected by the 6 cameras on our MAV. The corresponding value used by RoboRun varies between zero and 60000$m^{3}$ to account for the larger voxels used at low levels of precision. 

For the baseline, the map volume communicated to the planner and planned within (OctoMap to planner, and planner volume) is 150000 $m^{3}$, approximately matching the size of an average warehouse~\cite{avg_wh_size}. RoboRun can expand this value by up to 100x when low precision relaxes the deadline constraints, improving long-distance and multi-warehouse deliveries. 

The maximum velocity for both designs is chosen experimentally such that at least 80\% of flights are collision-free, given all of the previously set knob settings. This resulted in a maximum velocity of 6 $m/s$ for RoboRun and 0.5 $m/s$ for spatial oblivious designs.  

\begin{table}[]
\centering
\small
\begin{tabular}{|c|c|c|}
\hline
\textbf{Variable}                                                           & \textbf{Static} & \textbf{Dynamic} \\ \hline
Point cloud precision ($m$)                                                 & 0.3                   & \begin{tabular}[c]{@{}c@{}}{[}0.3 \ldots 9.6{]}\end{tabular}         \\ \hline
\begin{tabular}[c]{@{}c@{}}OctoMap to planner precision ($m$)\end{tabular} & 0.3                   & \begin{tabular}[c]{@{}c@{}}{[}0.3 \ldots 9.6{]}\end{tabular}         \\ \hline
OctoMap volume ($m^3$)                                                         & 46000                 & {[}0 \ldots 60000{]}         \\ \hline
\begin{tabular}[c]{@{}c@{}}OctoMap to planner volume ($m^3$)\end{tabular}    & 150000                & {[}0 \ldots 1000000{]}       \\ \hline
Planner volume ($m^3$)                                                         & 150000                & {[}0 \ldots 1000000{]}       \\ \hline
\end{tabular}
\caption{Knob values and ranges}
\label{fig:knob_setting_values}
\end{table}

\bibliographystyle{plain}
\bibliography{references}